\documentclass[lettersize,journal]{IEEEtran}
\usepackage{amsmath,amsfonts}
\usepackage{algorithmic}
\usepackage{algorithm}
\usepackage{array}
\usepackage[caption=false,font=normalsize,labelfont=sf,textfont=sf]{subfig}
\usepackage{textcomp}
\usepackage{stfloats}
\usepackage{url}
\usepackage{verbatim}
\usepackage{graphicx}
\usepackage{booktabs}
\usepackage{cite}
\usepackage{ragged2e}
\usepackage[justification=centering]{caption}
\usepackage{amsmath}
\usepackage[table]{xcolor}
\usepackage{orcidlink}
\usepackage{hyperref}
\hypersetup{hidelinks=true}
\usepackage{color}
\usepackage{multirow}

\newtheorem{remark}{Remark}

\begin{document}

\title{VisGeo-PipeTrack: A Visual Geometric Servo Control Framework with MPC-RCAPID Control for UAV-Based Dynamic Pipes Tracking}

\author{\IEEEauthorblockN{
        Zhenyang Gao~\orcidlink{0009-0002-7306-1958}
        ,
        \and
        Jianlin Chen~\orcidlink{0009-0002-7306-1958}
        ,
        \and
        Yang Yu~\orcidlink{0009-0002-7306-1958}
        ,
        \and 
        Honghao Shan~\orcidlink{0009-0009-4676-0279}
        ,
        \and 
        Jiale Gao~\orcidlink{0009-0008-7918-4477}
        and
        Ziyan Jia
    }
    \thanks{
    Z. Gao is with the School of Civil Aviation, Northwestern Polytechnical University, 127 West Youyi Road, Xi’an Shaanxi, 710072, P. R. China. Email:zhenyang\_gao@nwpu.edu.cn.}
    \thanks{
    J. Chen is with the School of Civil Aviation, Northwestern Polytechnical University, 127 West Youyi Road, Xi’an Shaanxi, 710072, P. R. China. Email:jianlin.chen@nwpu.edu.cn.}
    \thanks{
    Y. Yu is with the School of Aeronautics and Astronautics, Shanghai Jiao Tong University, Shanghai 200240, P. R. China. Email: yuyang22@sjtu.edu.cn.}
    \thanks{
    H. Shan is with the School of Civil Aviation, Northwestern Polytechnical University, 127 West Youyi Road, Xi’an Shaanxi, 710072, P. R. China. Email:honghao\_shan@mail.nwpu.edu.cn.}
    \thanks{
    J. Gao is with the School of Civil Aviation, Northwestern Polytechnical University, 127 West Youyi Road, Xi’an Shaanxi, 710072, P. R. China. Email:jiale.gao@mail.nwpu.edu.cn.}
    \thanks{
    Z. Jia is with the School of Civil Aviation, Northwestern Polytechnical University, 127 West Youyi Road, Xi’an Shaanxi, 710072, P. R. China. Email:jiaziyan@mail.nwpu.edu.cn.}
    }

\markboth{VisGeo-PipeTrack: A Visual Geometric Servo Control Framework with MPC-RCAPID Control for UAV-Based Dynamic Pipes Tracking}%
{Shell \MakeLowercase{\textit{et al.}}: VisGeo-PipeTrack: A Visual Geometric Servo Control Framework with MPC-RCAPID Control for UAV-Based Dynamic Pipes Tracking}

\maketitle

\begin{abstract}
While unmanned aerial vehicles (UAVs) are increasingly deployed for pipeline inspection, prevailing research predominantly assumes static pipes. In real-world industrial environments, pipes often exhibit dynamic movements, such as vibrations. This imposes more stringent requirements on UAVs' adaptive control and computational efficiency than in static scenarios. Therefore, this paper proposes a servo control framework named VisGeo-PipeTrack, which utilizes geometric constraints to enable UAV tracking of dynamic pipes. To achieve efficient UAV positioning, YOLOv5 is employed to detect labels on the pipe. The positions are derived from visual geometric constraints based on these detected labels to enable precise localization. To address tracking control challenges, a model predictive control based real-time corrected adaptive PID (MPC-RCAPID) controller is proposed. This controller employs the Recursive Least Squares method for real-time model estimation and integrates rolling horizon optimization to dynamically adjust PID parameters. Experiments conducted on both the AirSim simulator and the Tello UAV platform validate the effectiveness of the proposed framework. Simulation comparisons with standard PID, fractional-order PID, adaptive PID, and fuzzy PID demonstrate that MPC-RCAPID achieves the shortest settling time (3.26 s) and the lowest average tracking error (3.52\%). In physical experiments on the Tello platform, compared with standard PID control, MPC-RCAPID reduces the average tracking error by 55.52\%, 26.72\%, and 26.52\% across the three test dimensions. These results demonstrate that VisGeo-PipeTrack facilitates efficient autonomous tracking in dynamic pipe environments.

\end{abstract}


\section{Introduction}
\IEEEPARstart{U}{nmanned} Aerial Vehicles (UAVs), owing to their high maneuverability and compact size, have emerged as a core technology for autonomous inspection within confined spaces, including industrial pipes and ventilation ducts \cite{Ibrahim2025} \cite{Fadl2024}. They are primarily useful in overcoming the accessibility constraints faced by human personnel and ground-based robotic systems, thereby providing critical support for industrial inspection tasks.

Inspections within confined spaces present challenges intrinsic to such environments, including limited volume, poor lighting, and aerodynamic disturbances \cite{Papachristos2019} \cite{Ogai2018}. These characteristics impose stringent demands on UAV systems, particularly regarding positioning accuracy, attitude control robustness, and interference resistance  \cite{Dekai2026}. Beyond these requirements, confined operational spaces often present dynamic changes, such as vibration in industrial piping systems. It imposes additional demands on system stability and adaptive control. Such dynamic changes induce variations in the pipeline's position, thereby increasing the risk of in-flight collisions for the UAV\cite{GursoyDemir2025}. However, extant research lacks effective solutions for UAV flight within dynamic pipes. Targeted technological innovation is therefore imperative to achieve breakthroughs for UAV operations in these environments.

\begin{figure}
    \centering
    \includegraphics[width=1\linewidth]{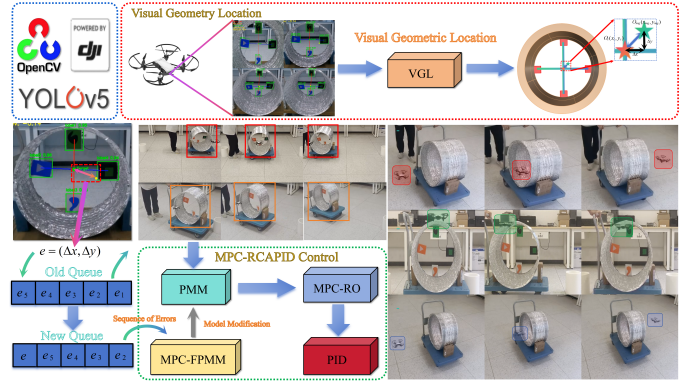}
    \caption{\justifying This paper proposes a servo control framework for the dynamic pipes tracking based on visual geometry named VisGeo-PipeTrack. The framework consists of a YOLOv5-based lightweight visual positioning servo system and an MPC-based real-time corrected adaptive PID control strategy, which includes three key components: MPC-FPMM (MPC Forward Prediction Model Modification), PMM (Prediction Model Modification), MPC-RO (MPC Rolling Optimization), and PID. This framework can enhance the performance of UAVs in tracking dynamic pipes.}
    \label{fig:1}
\end{figure}

Extensive research has developed effective solutions for UAV navigation within static pipeline environments. Current systems are typically designed with the assumption of fixed positions and lack mechanisms to respond to dynamic positional changes. Wang et al. proposed an autonomous flight system integrating virtual omnidirectional perception with airflow disturbance planning \cite{Wang2025}. Their system enables navigation in static tunnels as narrow as 0.5m in diameter, including those with complex cross-sections. However, it assumes fixed tunnel geometry and position, neglecting inherent tunnel motion. Zhao et al. developed an H-shaped tilt-rotor quadcopter with a disturbance observer and fixed-time sliding mode control for stable \cite{Zhao2024}. The quadcopter performs high-angle flights in static narrow passages. It is effective for fixed environments like rubble but lacks adaptation to passage displacement. Martin et al. used a robotic platform to characterize duct aerodynamics and achieved stable hovering in a static 35cm duct via neural network localization \cite{Martin2025}, yet their work does not account for dynamic effects like duct vibration or translation. Xie et al. employed deep reinforcement learning to traverse static gaps at tilt angles up to $60^\circ$, but the gap properties were predefined and unchanging\cite{Xie2023}. Chen et al. implemented a reinforcement learning strategy for real-time attitude adjustment in response to static gaps, with the dynamic response limited to the UAV's own pose \cite{Chen2023}. Wang et al. introduced a velocity planning method optimizing perception and aerodynamic stability in static tunnels, but the method does not incorporate mechanisms for tunnel dynamics\cite{Wang2022}. In summary, existing control strategies universally presuppose static environments. When dynamic motion occurs in pipes, current methods lack the capability to update positional references and control targets in real time. It results in accumulating state estimation errors and lagging control responses, which consequently elevate collision risk. This gap restricts the practical deployment of UAVs in dynamic confined spaces and thus underscores the necessity for an efficient control framework. 

To address the above challenges, this paper proposes  a servo control framework, named VisGeo-PipeTrack, for UAV-based dynamic pipes tracking based on visual geometry. The structure of the control framework is shown in Fig. \ref{fig:1}, which is characterized by two technical contributions:

\begin{enumerate}
    \item  A lightweight visual positioning system is realized through YOLOv5 label detection and geometric constraint resolution \cite{Han2024}. It employs YOLOv5 to detect labels attached to the pipe. The geometric intersections derived from these detected labels serve as target points for UAV tracking. This strategy ensures positioning accuracy while reducing computational overhead, thereby improving the UAV's dynamic response \cite{Liu2025}.
    \item  A model predictive control-based real-time corrected adaptive PID (MPC-RCAPID) control strategy. It consists of three integrated components: (1) Tracking error prediction through model predictive control (MPC); (2) Real-time model parameter estimation via the recursive least squares (RLS) algorithm \cite{Lai2024}; (3) Online PID parameter optimization in rolling time domain. This integration resolves limitations of standard PID controllers with fixed parameters and insufficient adaptability.
\end{enumerate}

The remainder of this paper is structured as follows. Section II introduces the basic technologies (YOLOv5 target detection algorithm, PID controller and dynamic pipes movement). Section III elaborates the pipe localization method based on visual geometric constraints. Section IV presents the design principle of the MPC-RCAPID  control strategy. In Section V, the performance of the proposed framework is verified by simulation and physical experiments. Section VI concludes the paper and looks forward to future research directions.

\section{Preliminaries}

\subsection{Problem Statement and Dynamic Pipes Movement}
This paper investigates the tracking of a UAV within a dynamic pipe environment. Existing UAV methods for pipe inspection are designed for static pipes and fail under dynamic pipe motion such as vibration or translation. When the pipe moves, the UAV cannot update its relative position or control target in real time, leading to error accumulation and increased collision risk. Therefore, this work solves the problem of enabling a UAV to robustly track a dynamically moving pipe using only onboard visual sensing and an adaptive control strategy.

\begin{figure}
    \centering
    \includegraphics[width=1\linewidth]{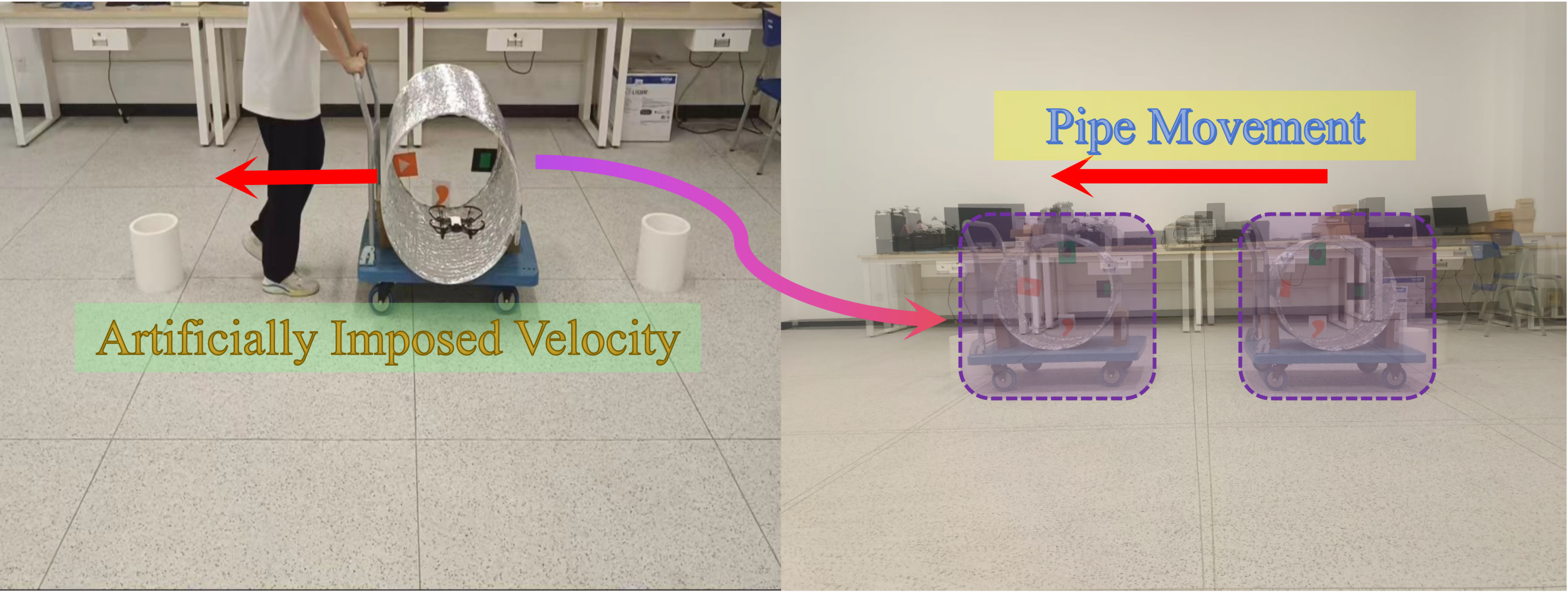}
    \caption{\justifying The application of variable speeds to the pipe induces positional changes, thereby simulating pipes vibrations under actual operating conditions.}
    \label{fig:3}
\end{figure}

In contrast to a static pipe, a dynamic pipe is characterized by variable movement speeds and positional offsets. The experimental protocol involves imposing a movement velocity on the pipe through human intervention. This procedure is implemented to ascertain whether the proposed framework can facilitate the UAV’s accurate and timely tracking of the pipe. The movement of the dynamic pipe is shown in Fig. \ref{fig:3}.

\subsection{YOLOv5 Target Detection Algorithm}
To meet the requirement of pipe positioning, the UAV must identify and detect pipe labels. In light of this necessity, the YOLOv5 target detection algorithm is selected for this task. The reason why is that YOLOv5 remains the most stable version that achieves a favorable trade-off between detection accuracy and processing speed on embedded platforms \cite{AJayasudha2025} \cite{Ganapathy2026}. Consequently, YOLOv5 is deemed the optimal choice for this application.
\subsection{PID Controller}
The PID controller, recognized as the most practical control algorithm in the field of  UAV control, consists of three components: the proportional term (P), the integral term (I), and the derivative term (D) \cite{Kiam2005}. By integrating these three control methods, the error of the controlled object can be effectively corrected. This controller ensures the stable operation of the control system and its transfer function as follows:
\begin{equation}
\label{deqn_ex1a}
\begin{array}{l}
G\left( s \right) = {K_P} + {K_I}\frac{1}{s} + {K_D}s\\
{\quad\quad}= {K_P}\left( {1 + \frac{1}{{{T_I}s}} + {T_D}s} \right)
\end{array}
\end{equation}
where $K_P$ represents proportional gain, $K_I$ represents integral gain, $K_D$ represents derivative gain, $T_I$ represents integral time constant, and $T_D$  represents derivative time constant.

\section{Dynamic Pipes Tracking Based on Visual Geometric Constraints}
The visual servo system is a fundamental component in pipe tracking tasks, which significantly affects the accuracy and real-time performance of tracking task \cite{Garcia2009}. In this section, a lightweight tracking servo system based on visual geometric constraints is designed to enhance the performance of pipes positioning.

\subsection{Label Visual Localization}
\begin{figure}
    \centering
    \includegraphics[width=1\linewidth]{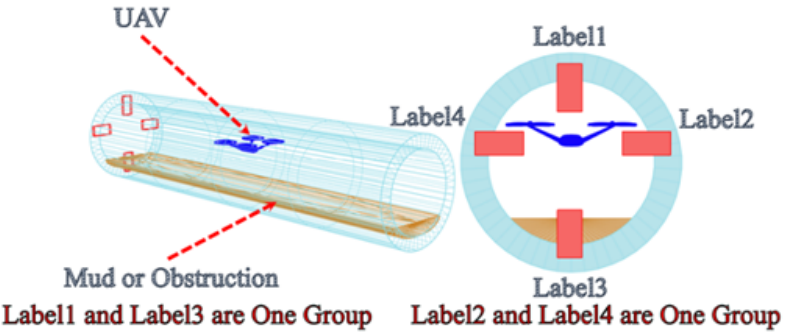}
    \caption{\justifying Guide the UAV to track the pipe and achieve relative positioning using the four labels attached to the pipe.}
    \label{fig:4}
\end{figure}

In this study, the YOLOv5 algorithm is employed to detect labels and construct the perception of the UAV within the pipe space. Four points are selected and marked on the pipe’s side wall, forming two groups of two points each. The two groups of points are each connected to form two lines. The intersection point of these two lines represents the relative position that the UAV maintains within the pipe. (In certain scenarios, the presence of oil stains or other obstacles within the pipe may impede the UAV’s flight, necessitating that the intersection point may not correspond to the pipe’s geometric center.) The positions of the four labels within the UAV camera’s field of view are detected using YOLOv5. Subsequent algorithms compute the UAV’s relative position relative to the pipe, thereby facilitating the UAV’s real-time perception of the pipe environment. The effect of label localization guiding the UAV to follow the pipe is illustrated in Fig. \ref{fig:4}.

\begin{figure}
    \centering
    \includegraphics[width=1\linewidth]{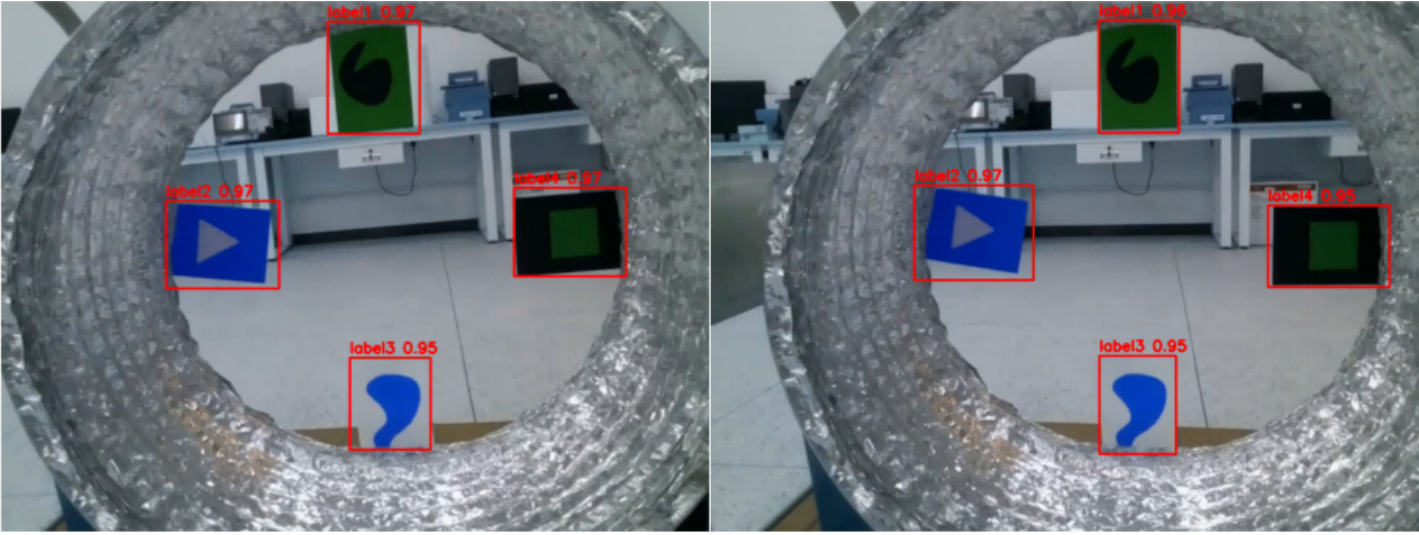}
    \caption{\justifying The YOLOv5 object detection algorithm identifies the positions of four labels for subsequent localization calculations.}
    \label{fig:5}
\end{figure}

In real experimental scenarios, the recognition confidence for the four labels consistently remains above 90\% across varying lighting conditions, flight attitudes, and label distances which is shown in Fig. \ref{fig:5}. Additionally, the frame rate of the UAV’s video streaming is sustained at over 25 frames per second. It satisfies the requirements for rapid and precise perception by the UAV of the pipe environment. 
Upon completion of labels recognition by YOLOv5, the visual positions of the four labels on the UAV camera are outputted as ${P_1}\left( {{x_1},{y_1}} \right)$, ${P_2}\left( {{x_2},{y_2}} \right)$, ${P_3}\left( {{x_3},{y_3}} \right)$ and ${P_4}\left( {{x_4},{y_4}} \right)$ respectively. These four positions serve as inputs to the subsequent localization algorithm.

\begin{remark}
The method of identifying an intersection point from two groups of lines is used to improve compatibility with various pipe structures. In certain engineering scenarios, the UAV is not required to maintain flight precisely at the pipe's geometric center but must hold a specific relative position. Therefore, a stable flight position for the UAV can be defined by configuring the locations of labels. The proposed visual localization method relies on these pre-attached labels, which is a deliberate design choice to ensure reliable detection under varying lighting and viewing angles. Additionally, to enhance geometric localization accuracy in long pipeline environments, labeled images captured at varying distances can be used to augment the YOLOv5 training set. For long pipelines without pre-installed labels, the framework can be extended by using natural features. Such extensions are part of our future work.
\end{remark}

\subsection{Geometric Location Solution of Pipes}
\begin{figure}
    \centering
    \includegraphics[width=0.7\linewidth]{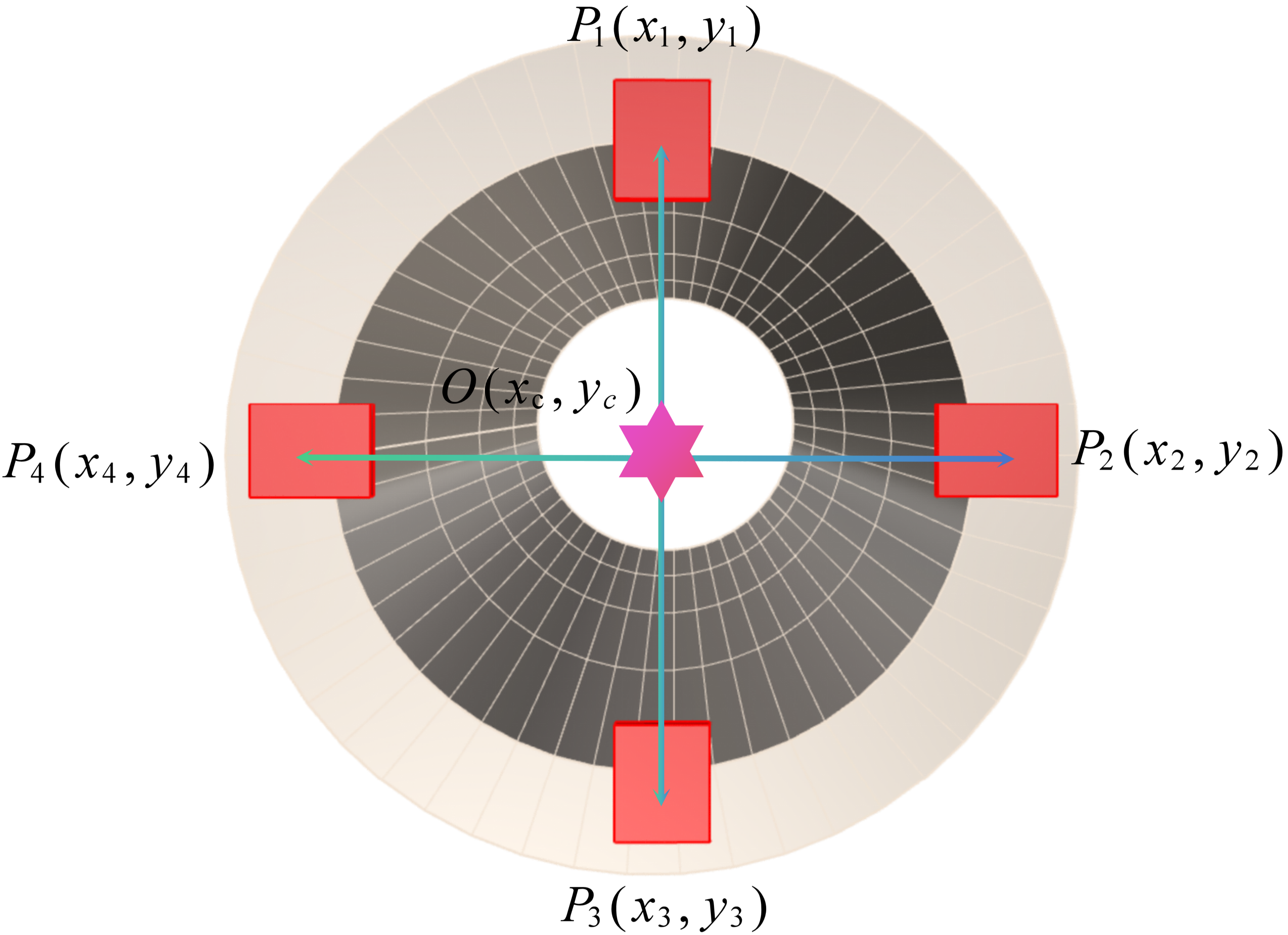}
    \caption{\justifying The intersection point of the two groups of intersecting lines indicates the position where the UAV must remain inside the pipe.}
    \label{fig:6}
\end{figure}

\begin{figure}
    \centering
    \includegraphics[width=1\linewidth]{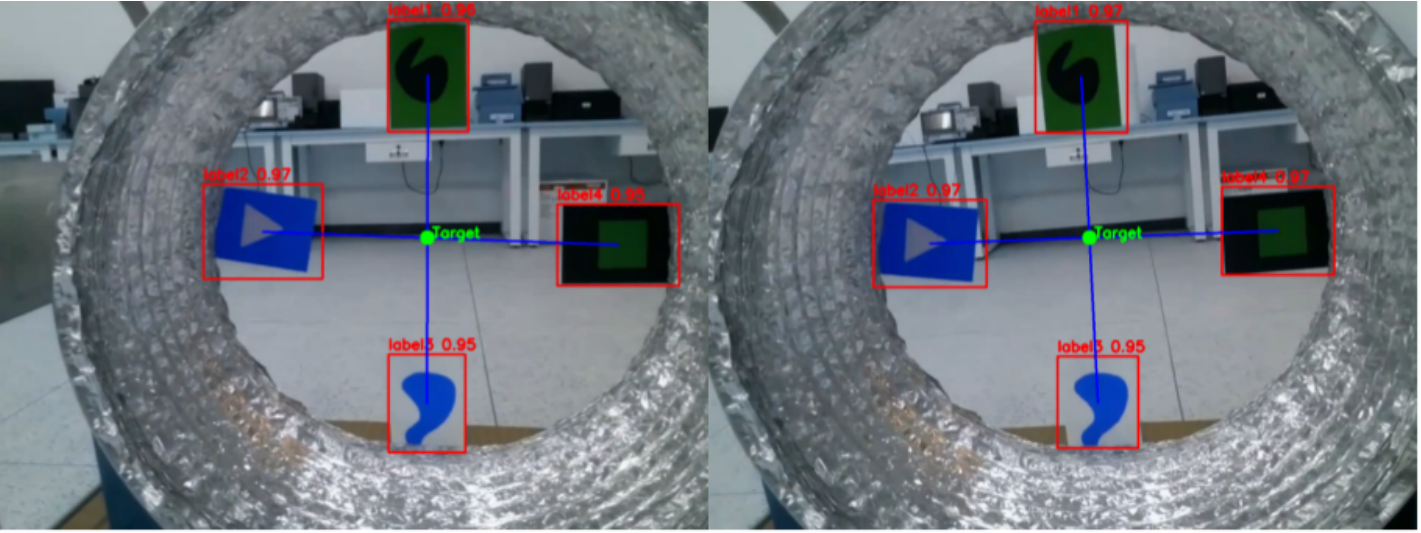}
    \caption{\justifying By combining the positions of the four labels detected by YOLOv5, the intersection point where the UAV needs to stabilize inside the pipe can be calculated.}
    \label{fig:7}
\end{figure}

The dynamic pipes tracking servo system is centered on computing the pipe’s position within an image. This process relies on geometric constraints. The position is compared to the visual center of the UAV’s camera and relative position of the UAV within the pipe is sensed by calculating the error. The intersection  of $O\left( {{x_c},{y_c}} \right)$ the two intersecting lines ${P_1} - {P_3}$ and ${P_2} - {P_4}$ is the relative position needed to keep the UAV inside the pipe. The intersection position is shown in Fig. \ref{fig:6}.
The intersection point of two intersecting lines in a visual image can be derived by parametric equations. The parametric equation of the line $L_1$(connecting $P_1$ and $P_3$) is given by:
\begin{equation}
\label{eq2}
{L_1} = \left\{ \begin{array}{l}
x = {x_1} + w\left( {{x_3} - {x_1}} \right)\\
y = {y_1} + w\left( {{y_3} - {y_1}} \right)
\end{array} \right.{\rm{     }}\left( {w \in \left[ {0,1} \right]} \right)
\end{equation}
Similarly, the parametric equation of line $L_2$ (connecting $P_2$ and $P_4$ ) is shown as:
\begin{equation}
\label{eq3}
{L_1} = \left\{ \begin{array}{l}
x = {x_2} + v\left( {{x_4} - {x_2}} \right)\\
y = {y_2} + v\left( {{y_4} - {y_2}} \right)
\end{array} \right.{\rm{     }}\left( {v \in \left[ {0,1} \right]} \right)
\end{equation}
where $w$ and $u$ serve as the parameters of the line parametric equation, employed to describe the position of a point on the line. From equation (\ref{eq2}) and equation (\ref{eq3}), the following matrix equation can be derived:
\begin{equation}
\label{deqn_ex1a}
\left[ {\begin{array}{*{20}{c}}
{{x_3} - {x_1}}&{ - ({x_4} - {x_2})}\\
{{y_3} - {y_1}}&{ - ({y_4} - {y_2})}
\end{array}} \right]\left[ {\begin{array}{*{20}{c}}
w\\
v
\end{array}} \right] = \left[ {\begin{array}{*{20}{c}}
{{x_2} - {x_1}}\\
{{y_2} - {y_1}}
\end{array}} \right]
\end{equation}
By solving the matrix equation, we have:
\begin{equation}
\label{deqn_ex1a}
w = \frac{{\left[ {\begin{array}{*{20}{c}}
{{x_2} - {x_1}}&{{x_4} - {x_2}}\\
{{y_2} - {y_1}}&{{y_4} - {y_2}}
\end{array}} \right]}}{{\left[ {\begin{array}{*{20}{c}}
{{x_3} - {x_1}}&{{x_4} - {x_2}}\\
{{y_3} - {y_1}}&{{y_4} - {y_2}}
\end{array}} \right]{\rm{ }}}}{\rm{ \quad }}v = \frac{{\left[ {\begin{array}{*{20}{c}}
{{x_3} - {x_1}}&{{x_2} - {x_1}}\\
{{y_3} - {y_1}}&{{y_2} - {y_1}}
\end{array}} \right]}}{{\left[ {\begin{array}{*{20}{c}}
{{x_3} - {x_1}}&{{x_4} - {x_2}}\\
{{y_3} - {y_1}}&{{y_4} - {y_2}}
\end{array}} \right]{\rm{ }}}}
\end{equation}
Then, the intersection point of the two lines is denoted by:
\begin{equation}
\label{deqn_ex1a}
\left\{ \begin{array}{l}
{x_c} = {x_1} + w\left( {{x_3} - {x_1}} \right)\\
{y_c} = {y_1} + w\left( {{y_3} - {y_1}} \right)
\end{array} \right.{\rm{   or  \quad}}\left\{ \begin{array}{l}
{x_c} = {x_2} + v\left( {{x_4} - {x_2}} \right)\\
{y_c} = {y_2} + v\left( {{y_4} - {y_2}} \right)
\end{array} \right.
\end{equation}
The computed effect of the intersection of two intersecting lines is shown in Fig. \ref{fig:7}.

After obtaining the intersection position and comparing it with the vision center ${O_{img}}\left( {{x_{img}},{y_{img}}} \right)$ of the UAV, the position error can be obtained, which contains the direction and distance. Error information can be used to guide the UAV in adjusting its flight attitude.
\begin{equation}
\label{deqn_ex1a}
e = \left( {{x_c} - {x_{img}},{y_c} - {y_{img}}} \right)
\end{equation}

\begin{remark}
In our experiments, the UAV's yaw angle relative to the pipe axis is maintained within ±15° by the low-level attitude controller. Within this range, perspective distortion of the four labels is limited, and the computed intersection point remains a sufficiently accurate reference for visual servoing. Therefore, no explicit compensation for yaw is required in the proposed framework.
\end{remark}

\section{MPC Based Real-Time Corrected Adaptive PID}
In dynamic operating environments with vibration or non-uniform motion, standard fixed-parameter PID controllers fail to handle sudden speed variations and noise interference effectively, often causing tracking error overshoot or target loss. In this subsection, a composite control strategy termed MPC-RCAPID is proposed, which integrates the rolling optimization principle of MPC, a real-time feedback correction mechanism, and an adaptive PID controller. The core logic of the control strategy is that the visual servoing system performs the real-time data sensing. Then, the error corrected prediction model generates future multiple steps error sequences.  Subsequently, optimal PID parameters are derived through rolling optimization, which produces UAV tracking control quantities. By integrating MPC’s forward prediction of pipes dynamics and ensuring engineering feasibility via adaptive PID tuning, the strategy is suited for time varying servo systems such as UAV pipes tracking.

\begin{figure}
    \centering
    \includegraphics[width=1\linewidth]{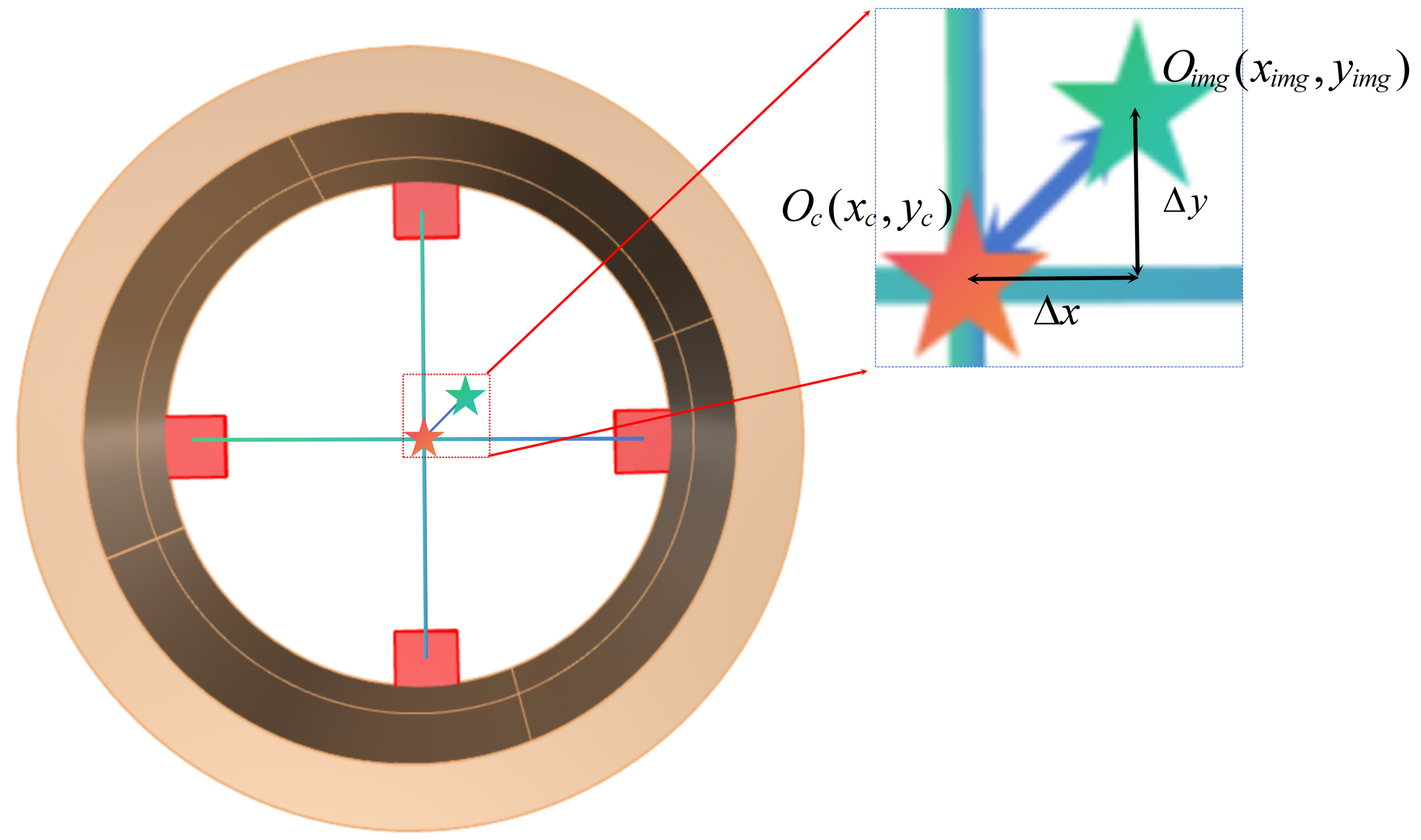}
    \caption{\justifying The position error between the UAV's current location and the target location. This error is used for subsequent adjustments to the UAV's flight attitude.}
    \label{fig:8}
\end{figure}

\subsection{Modification of the MPC Forward Prediction Model}
The MPC framework is designed to predict the system’s future state. The prediction of position error variation is achieved through a dedicated model. This model is subsequently adjusted in real-time, which enhances its prediction accuracy\cite{Zhan2025}. Under the assumption that the error variation is governed by a second order function model, the predicted error for future time $t + k$ (with $k=1,2,3,\dots,N$) is defined based on the current position error $e(t)$:

\begin{equation}
\label{eq:pred}
\hat e(t + k) = e(t) + A \cdot k + B \cdot k^2
\end{equation}
where $\hat e(t + k)$ denotes the predicted error at time $t + k$, $e(t)$ represents the current position error at time $t$, $A$ represents the error influence due to the speed of error change, and $B$ represents the error influence due to the acceleration of error change. When the system advances to time $t + k$, the actual measured error $e_{\text{real}}(t+k)$ becomes available. The discrepancy between the actual measurement error and the predicted error is denoted as:
\begin{equation}
\label{eq:delta}
\delta(k) = e_{\text{real}}(t + k) - \hat e(t + k)
\end{equation}

\begin{remark}
Note that \(\delta(k)\) is computed only after the system reaches time \(t+k\) and it is not known at the current time \(t\). This \(\delta(k)\) is then used to correct the predictions for the remaining future steps.
\end{remark}
To dynamically adjust and compensate for future predictions, the predictions for the remaining future steps are corrected by a decaying correction as follows:
\begin{equation}
\label{eq:corr}
\hat{e}_{\text{new}}(t+k+i) = \hat{e}(t+k+i) + \gamma(k) \cdot \beta_i \cdot \delta(k)
\end{equation}
where $i = 1,2,\dots,N-k$ denotes the number of steps after time $t+k$, and $\hat{e}_{\text{new}}(t+k+i)$ represents the predicted error after adjustment and updating. The error transfer coefficient $\beta_i = e^{-i}$ ensures that the correction strength decays exponentially with the step size, avoiding excessive intervention. The introduced adaptive correction factor $\gamma(k)$ adjusts the correction intensity according to the magnitude of $\delta(k)$ to trade off response speed and stability, and is defined as:
\begin{equation}
\label{eq:gamma}
\gamma(k) = \min \left\{ \gamma_{\max},\; \max \left\{ \gamma_{\min},\; \frac{|\delta(k)|}{\delta_{\text{ref}}} \right\} \right\}
\end{equation}
where $\gamma_{\max}$ represents the upper limit of the correction strength to avoid excessive correction, $\gamma_{\min}$ represents the lower limit to ensure a baseline correction even when the error is minimal, and $\delta_{\text{ref}}$ is a reference value for normalization that maps errors of varying magnitudes into a unified correction intensity range.
\begin{remark}
The selection of hyperparameters (such as $\gamma_{\max}$, $\gamma_{\min}$, $\delta_{\text{ref}}$) requires adjustment based on the control system and experimental results.
\end{remark}

Furthermore, the acceleration of error change serves as a crucial parameter in the error prediction model, and the accuracy of its calculation directly influences the precision of error prediction. Based on the uncertainty associated with error change, the RLS algorithm is employed to estimate the acceleration and speed of error change in real time using historical error data. This approach enables the long-term self-adaptation of the model parameters. For the problem of estimating the varying error acceleration, the performance index over past measurements is:

\begin{equation}
\label{eq:J_rls}
J_{\text{RLS}} = \sum_{k=1}^{N} \lambda_{\text{RLS}}^{N-k} \left( \delta(k) \right)^2
\end{equation}
where \(J_{\text{RLS}}\) is the performance index for RLS parameter estimation, \(\delta(k) = e_{\text{real}}(k) - \hat{e}(k)\) is the prediction error at historical step \(k\), and the index \(k\) in this RLS formulation indexes historical time steps (past data), which is independent of the future-step index \(k\) used in the prediction model above. The forgetting factor \(\lambda_{\text{RLS}} \in (0,1]\) assigns higher weights to recent errors, prioritizing the latest data to correct model biases and enhance prediction accuracy.

To minimize the performance index \(J_{\text{RLS}}\), the derivative of the optimal solution \([A, B]\) for the predictive model parameters is computed and set to zero, resulting in a recursive formula for updating these parameters~\cite{Li2022}. Employing this formula requires calculating the RLS gain:

\begin{equation}
\label{eq:K}
K(k) = \frac{P(k-1) \cdot \phi(k)}{\lambda_{\text{RLS}} + \phi^T(k) \cdot P(k-1) \cdot \phi(k)}
\end{equation}
where \(\phi(k) = [k, k^2]\) is the regression vector, and \(P(k-1)\) is the covariance matrix at recursion step \(k-1\), which quantifies the uncertainty in parameter estimation. Higher uncertainty leads to an increase in the RLS gain, making the impact of new data more pronounced. The estimates of model parameters are updated as:

\begin{equation}
\label{eq:update}
\begin{array}{l}
\begin{bmatrix} \hat{A}(k) \\ \hat{B}(k) \end{bmatrix} = 
\begin{bmatrix} \hat{A}(k-1) \\ \hat{B}(k-1) \end{bmatrix} \\
\quad\quad\quad\quad + K(k) \left( e_{\text{real}}(k) - \phi^T(k) \begin{bmatrix} \hat{A}(k-1) \\ \hat{B}(k-1) \end{bmatrix} \right)
\end{array}
\end{equation}
When the estimation error is substantial (small), the amplitude of parameter updates is increased (reduced), allowing the estimation error to gradually converge. This process aids the UAV in accurately capturing the error change dynamics. The covariance matrix is updated by:

\begin{equation}
\label{eq:P}
P(k) = \frac{1}{\lambda_{\text{RLS}}} \left[ P(k-1) - K(k) \phi^T(k) P(k-1) \right]
\end{equation}
The covariance matrix converges as estimation accuracy improves, preventing parameter oscillation induced by an excessively large gain. As the estimation accuracy increases, \(P(k)\) and \(K(k)\) both decrease, making the algorithm gradually stable.
\begin{remark}
The RLS algorithm minimizes the sum of squared measurement and model prediction errors, thereby reducing the prediction error. Upon a sudden change in the pipe’s motion state, the adjustment range of the model parameters $[A,B]$ is widened to align with the actual motion state. Such adjustments are implemented to preserve the accuracy and stability of the error prediction model.
\end{remark}

\subsection{PID Tuning Based on MPC Rolling Optimization}
MPC achieves the optimal control law by solving an optimization problem within a finite time horizon. In the subsection, the idea of MPC is employed to tune the PID parameters to minimize the cumulative sum of squared tracking errors over the next $N$ steps. The UAV employs an adaptive PID algorithm with rolling optimization for dynamic pipes tracking control. The discrete PID controller of UAV is designed as:

\begin{equation}
\label{eq:pid}
\begin{aligned}
u(k) = & \; K_P(k) \cdot e(k) + K_I(k) \cdot \sum_{i=1}^k e(i) \\
       & + K_D(k) \cdot \left[ e(k) - e(k-1) \right]
\end{aligned}
\end{equation}
where $K_P$ represents the proportional control coefficient, $K_I$ represents the integral control coefficient, and $K_D$ represents the differential control coefficient. The cumulative sum of squared tracking errors for the subsequent $N$ steps is defined as:
\begin{equation}
\label{eq:J_mpc}
J_{\text{MPC}} = \sum_{k=1}^N \hat e_k^2 + \varphi_u \sum_{k=1}^N \Delta u_k^2
\end{equation}
where \(J_{\text{MPC}}\) is the cost function for MPC-based PID tuning, \(\hat e_k\) is the predicted tracking error at the \(k\)-th future step, \(\Delta u_k = u_k - u_{k-1}\) is the change rate of the control quantity, and \(\varphi_u\) is the penalty coefficient to trade off tracking accuracy and control smoothness. The gradient descent method is used to solve the PID parameters that minimize \(J_{\text{MPC}}\). For each parameter \(\theta = \{K_P, K_I, K_D\}\) to be optimized, the optimized parameters meet the following requirements:
\begin{equation}
\label{eq:grad}
\nabla J_{\text{MPC}}(\theta) = \frac{J_{\text{MPC}}(\theta + \Delta) - J_{\text{MPC}}(\theta - \Delta)}{2 \cdot \Delta}
\end{equation}
where $\Delta$ represents the step size of finite difference. The derivative of the parameters with respect to the performance index is approximated using the finite difference method~\cite{de2021}. This approach captures both the degree and direction of influence that changes in PID parameters have on $J_{\text{MPC}}$. Finally, the parameters are adjusted based on the negative direction of the gradient:

\begin{equation}
\label{eq:theta}
\theta(k + 1) = \theta(k) - \eta \cdot \nabla J_{\text{MPC}}(\theta(k))
\end{equation}
where $\eta$ represents the learning rate to regulate the magnitude of parameter updates. The proposed MPC-RCAPID controller enables the UAV to quickly and accurately track the dynamic pipe.

\begin{remark}
The MPC-RCAPID control algorithm incorporates MPC rolling optimization, a real-time correction mechanism, and adaptive PID control. The RLS algorithm is utilized to adjust the parameters of the predictive model in real-time. Its adjustment is combined with real-time error to correct prediction errors. In comparison to traditional PID control algorithms, the MPC-RCAPID approach is capable of swiftly responding to sudden changes in pipe motion states, thereby minimizing the tracking error of UAVs.
\end{remark}

\section{Simulation and Experimental Results}
\subsection{Tello UAV Platform}
\begin{figure}
    \centering
    \includegraphics[width=1\linewidth]{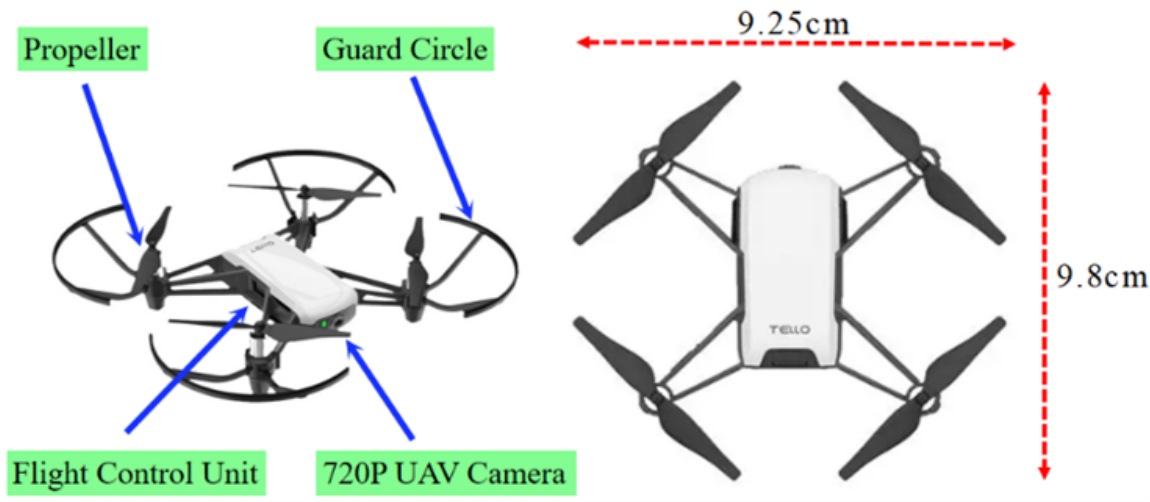}
    \caption{\justifying The Tello UAV platform used in experiments. Key components include the propeller, guard circle, flight control unit, and 720P onboard camera. The UAV dimensions are 9.8 cm (length) × 9.25 cm (width).}
    \label{fig:15}
\end{figure}

In the present investigation, the incorporation of visual servo control is deemed crucial for navigation and guidance purposes. This is due to the necessity for the UAV to autonomously track pipelines with precise geometric center localization. Furthermore, given the constraints of limited space and the dynamic uncertainties inherent in the pipeline environment, the UAV must possess a compact form factor and rapid responsiveness to ensure agile maneuverability within the pipeline. Consequently, the Tello UAV platform, with its compact size, is selected for experimental validation and system development. The Tello UAV platform is shown in Fig. \ref{fig:15}.

\subsection{Simulation Results Based on Airsim}
\begin{figure}
    \centering
    \includegraphics[width=1\linewidth]{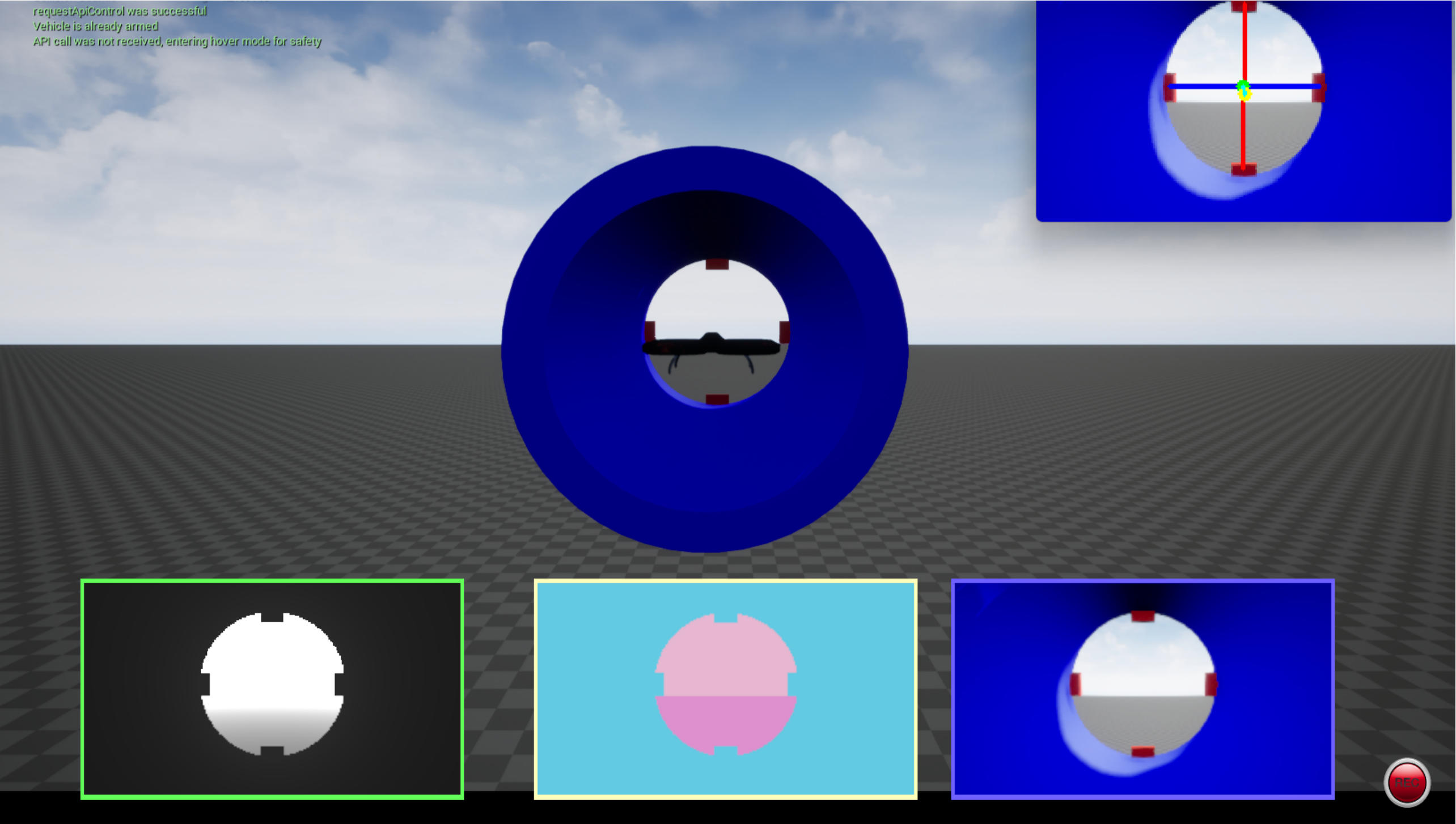}
    \caption{\justifying Simulation of UAV pipe tracking based on the Airsim and UE4 platforms, comparing the performance of the proposed method with the baseline method.}
    \label{fig:9}
\end{figure}

\begin{figure}
    \centering
    \includegraphics[width=1\linewidth]{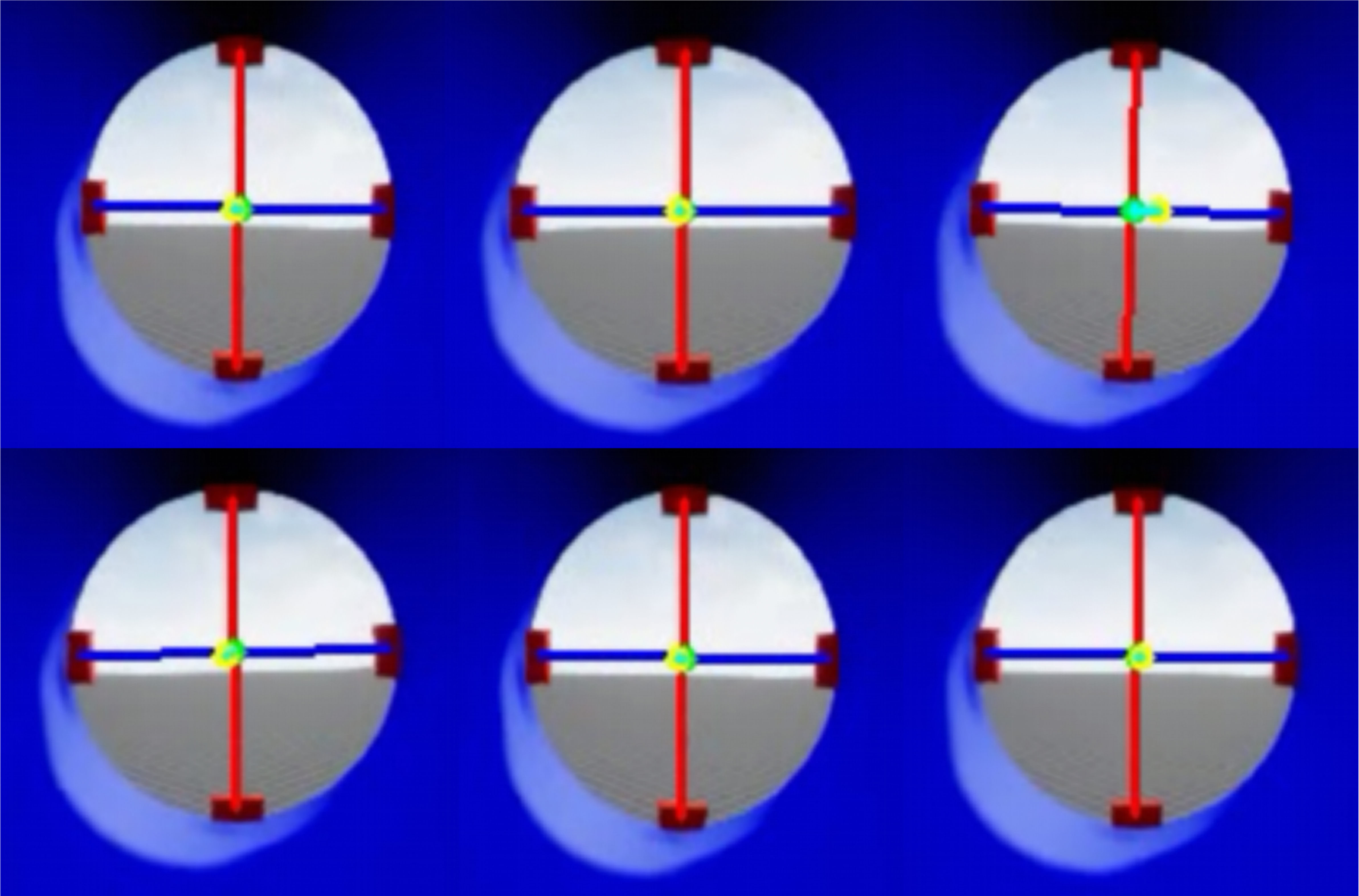}
    \caption{\justifying As shown in the figure, the positions of the UAV’s visual center and the intersection of the two label pairs during tracking are illustrated. The green point denotes the intersection of the two label pairs, and the yellow point denotes the UAV’s visual center. It can be observed that the visual center remains consistently close to the intersection, indicating good tracking performance.}
    \label{fig:10}
\end{figure}

To verify the trajectory tracking performance of the proposed algorithm in dynamic pipes scenarios, a simulation experiment is conducted using the AirSim platform. A quadrotor model with specific dimensions (0.18 m in $x$ direction, 0.11 m in $y$ direction, and 0.04 m in $z$ direction) is adopted for platform representation. The experimental environment is configured as a cylindrical pipe with an inner diameter of 0.4 m and a length of 1 m. The pipe moves at a constant velocity of 30 cm/s horizontally for 2 seconds, then stops and remains stationary to allow the UAV to make adjustments. This process simulates dynamic pipe constraints. Its primary objective is to achieve dynamic tracking control while maintaining a safe distance from the pipe wall and avoiding collisions. Two control strategies, PID and MPC-RCAPID, are implemented for comparative performance evaluation. The tracking effectiveness is evaluated during the entire operation period. The simulation effect of AirSim is shown in Fig. \ref{fig:9} and Fig. \ref{fig:10}.

\begin{remark}
The pipe movement speed in our experiments is set to 30 cm/s (300 mm/s), which is more than ten times higher than the typical industrial screening criterion for piping vibration (25 mm/s peak velocity as recommended by Fauske and Associates \cite{Conzen2017}). This deliberately extreme condition is chosen to validate the robustness of the proposed controller under dynamic disturbances that far exceed normal expectations. By demonstrating stable tracking at such a severe level, the framework is shown to effectively cover virtually all possible pipe vibration amplitudes encountered in real-world industrial scenarios.
\end{remark}

\begin{table}
    \centering
    \renewcommand{\arraystretch}{1.5}
    \caption{Comparison of Simulation Results}
    \label{tab:sim_comparison}
    \begin{tabular}{>{\bfseries}c
                   >{\centering\arraybackslash}m{1.8cm}  
                   >{\centering\arraybackslash}m{1.6cm}  
                   >{\centering\arraybackslash}m{1.6cm}  
                   >{\centering\arraybackslash}m{1.8cm}} 
    \toprule
    \textbf{Categories} & \textbf{Adjustment Time} & \textbf{Max Error} & \textbf{Average Error} \\
    \midrule
    MPC-RCAPID & 3.26s & 10.91\% & 3.52\% \\
    PID & 5.16s & 14.19\% & 6.96\% \\
    FO-PID & 8.14s & 13.14\% & 7.44\% \\
    A-PID& 13.62s & 16.40\% & 6.61\% \\
    F-PID & 11.83s & 16.32\% & 6.36\% \\
    \bottomrule
    \end{tabular}
    \begin{remark}
\textbf{Adjustment Time} represents the time from pipe movement to the last instance where the error exceeded the 10\% threshold. \textbf{Max Error} represents the maximum error during the adjustment time. \textbf{Average Error} represents the average error during the adjustment time.
\end{remark}
\end{table}

Table \ref{tab:sim_comparison} summarizes the control performance of all five controllers in the AirSim simulation environment. These methods include:a
\begin{enumerate}
    \item \textbf{MPC-RCAPID:} The proposed controller. It is used to validate the adaptive advantages in dynamic pipe tracking.
    \item \textbf{PID:} The classical PID controller. It is used as a performance baseline for comparison.
    \item \textbf{FO-PID:} The fractional-order PID controller. It is used to compare the robustness of fractional-order control\cite{Abro2024}.
    \item \textbf{A-PID:} The gain-scheduled adaptive PID controller. It is used to examine the effectiveness of error-driven simple adaptation\cite{Arfa2024}.
    \item \textbf{F-PID:} The fuzzy PID controller. It is used to compare knowledge‑based rule tuning with numerical optimization strategies\cite{Anh2025}.
\end{enumerate}
The results indicate a clear performance advantage of the proposed MPC-RCAPID framework over the traditional PID controller as well as other PID variants (FO-PID, A-PID, and F-PID). By predicting the motion state of the pipeline disturbance, the MPC-RCAPID framework enables adaptive online tuning of the PID controller parameters. This capability allows it to achieve the shortest adjustment time (3.26 s), the lowest maximum error (10.91\%), and the smallest average error (3.52\%) under lateral pipeline motion. In comparison, the standard PID controller requires an adjustment time of 5.16 s, with a maximum error of 14.19\% and an average error of 6.96\%. Among the other PID variants, FO-PID shows a longer adjustment time (8.14 s) and higher average error (7.44\%); A-PID and F-PID exhibit even longer adjustment times (13.62 s and 11.83 s, respectively) and larger maximum errors (16.40\% and 16.32\%). These results demonstrate that the proposed MPC-RCAPID framework effectively mitigates the key limitations of traditional PID and its common variants, including slow response and larger tracking errors, in dynamic pipeline tracking scenarios.

\subsection{Experimental Results Based on Tello}
\begin{figure}
    \centering
    \includegraphics[width=1\linewidth]{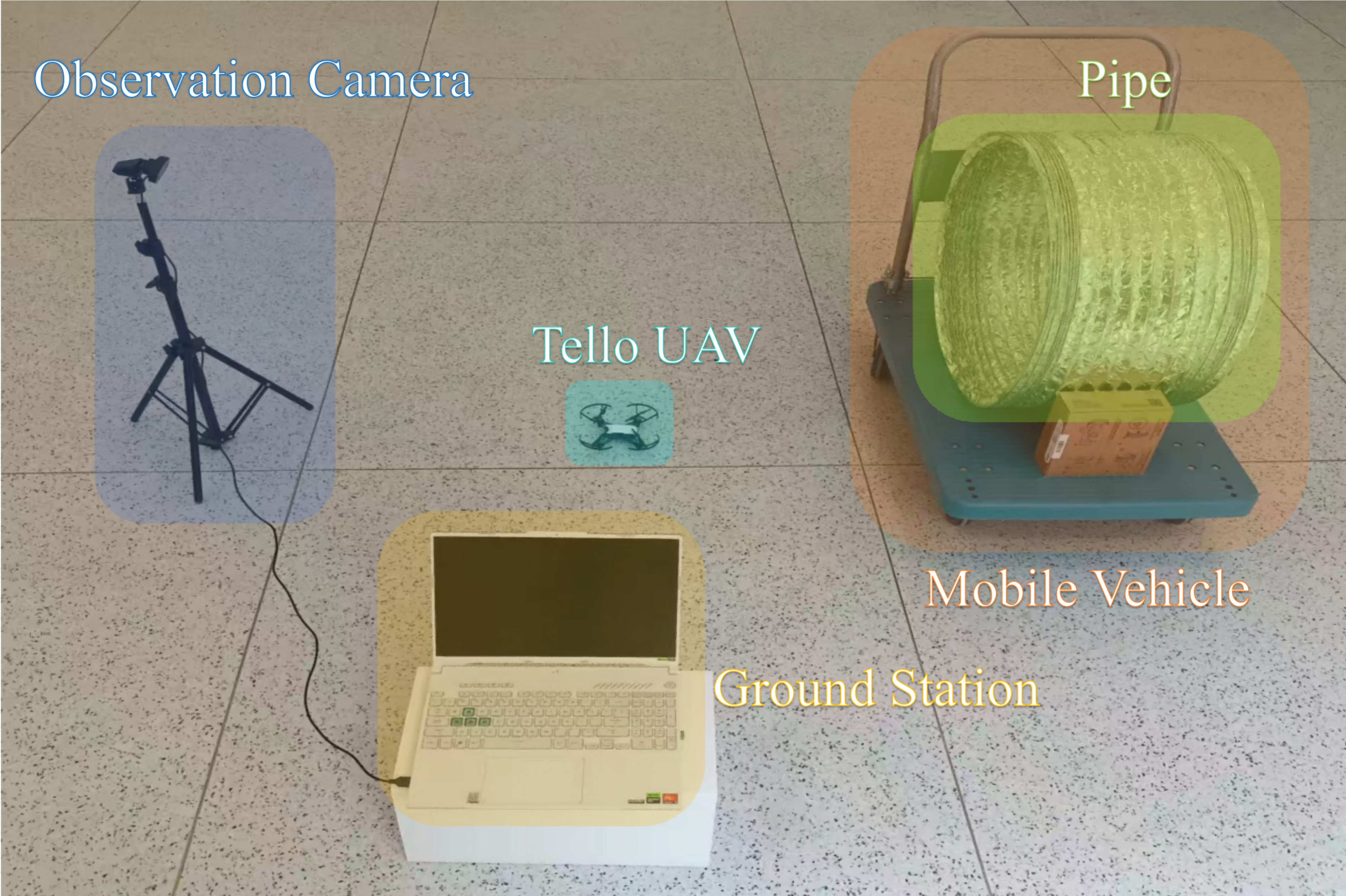}
    \caption{\justifying The experimental setup includes the observation camera, the Tello UAV  platform, the ground station, pipe, and the vehicle for moving the pipe.}
    \label{fig:11}
\end{figure}

To demonstrate the portability and real-world effectiveness of the proposed framework, an experimental platform is established for dynamic tracking outside and dynamic crossing inside pipes scenarios. The platform comprises core components including a Tello UAV, visual monitoring camera, ground station computer, physical pipes, and mobile vehicle. In the experimental process, initial control and task instructions are transmitted to the UAV via the UDP communication protocol~\cite{Tie2014}. Simultaneously, environmental image data collected by the UAV camera is continuously processed at the local terminal, including pipe labels detection, geometric positioning, and environmental perception. Sensing results are converted into real-time control instructions, which are transmitted back to the UAV through communication links, achieving closed-loop control and dynamic adjustment of UAV motion states. We compare the MPC-RCAPID controller with the standard PID controller in multi-scenario physical experiments to verify that the proposed algorithm not only achieves superior tracking performance consistent with simulation results on a real UAV, but also possesses good hardware portability. The experimental environment layout and parameter configurations refer to Fig. \ref{fig:11} and Table \ref{tab:parameters}.

\begin{table}
    \centering
    \renewcommand{\arraystretch}{1.5} 
    \caption{Experimental Parameters}
    \begin{tabular}{>{\bfseries}c>{\centering\arraybackslash}p{4cm}}
    \toprule
         \textbf{Categories} & \textbf{Parameters} \\
    \midrule
         Tello UAV Size & 9.8cm $\times$ 9.25cm $\times$ 4.1cm \\
         The Pipe Radius & 60cm \\
         The Pipe Length & 70cm \\
         The Pipe Movement Speed & 30±5cm/s \\
         UAV Camera Pixels & 720P \\
         Experimental Site Width & 320cm \\
         Experimental Site Length & 240cm \\
         Prediction Horizon $N$ & 5 steps\\
         Sampling Time $\Delta t$ &  0.05s \\
         Correction Decay Factor $\lambda_\beta$ & 0.9 \\
         RLS Forgetting Factor $\lambda_{\text{RLS}}$ & 0.98 \\
         Initial PID Gains $K_{P_0}, K_{I_0}, K_{D_0}$ & 20, 3.5, 6.5 \\     
    \bottomrule
    \end{tabular}
    \label{tab:parameters}
    \begin{remark}
    Experimental environment parameters are derived from real-world experimental settings. Experimental algorithm parameters are selected through a combination of simulation analysis and experimental testing. $N = 5$ trade-off computational load and predictive capability; $\Delta t = 0.05$ trade-off sampling frequency and computational load; $\lambda_\beta = 0.9$ ensures moderate correction strength for long-term predictions; $\lambda_{\text{RLS}} = 0.98$ adjusts the influence of different historical data on model predictions; $K_{p0}, K_{i0}, K_{d0} = 20, 3.5, 6.5$ represent initial PID parameters tuned through Airsim simulations and real-machine testing.
    \end{remark}

\end{table}

\begin{figure}
    \centering
    \includegraphics[width=1\linewidth]{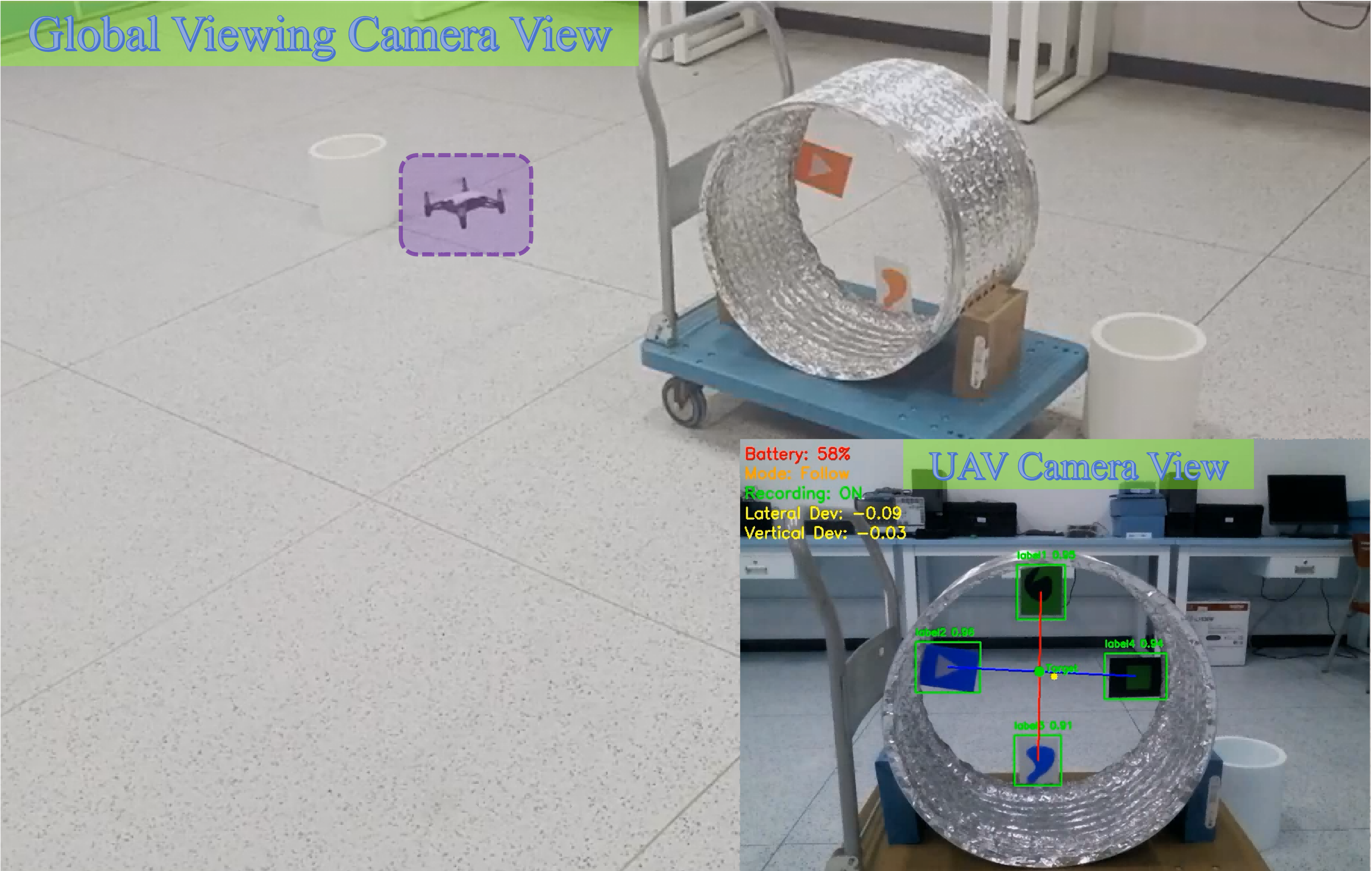}
    \caption{\justifying The experimental process can be observed through both the global viewing and the UAV camera viewing to monitor the UAV's flight attitude.}
    \label{fig:12}
\end{figure}

During the experiment, the UAV’s attitude is dynamically adjusted based on real-time control instructions from the MPC-RCAPID. This adjustment ensures stable relative positioning within the pipe. The ground station continuously displays real-time relative position data between the UAV and pipe, which provides a quantitative basis for evaluating the adjustment effect. The global view and ground station view are shown in Fig. \ref{fig:12}.

\begin{figure}
    \centering
    \includegraphics[width=1\linewidth]{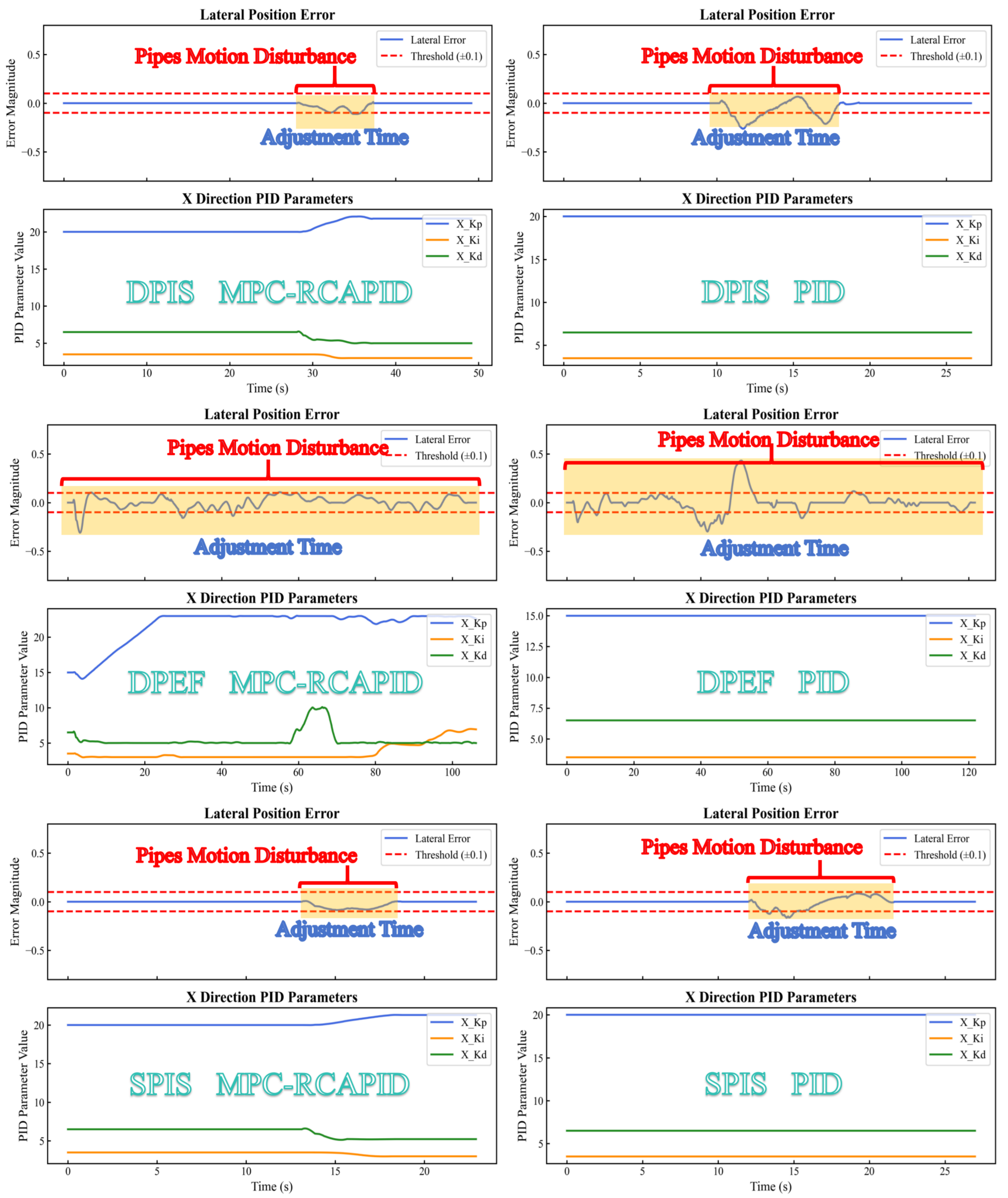}
    \caption{\justifying  Experimental comparison of the proposed MPC-RCAPID controller against the standard PID controller under three testing scenarios: DPIS (Dynamic Pipes Internal Shuttle), DPEF (Dynamic Pipes External Following), and SPIS (Static Pipes Internal Shuttle). In DPIS and SPIS, the pipe undergoes a brief motion disturbance then remains stationary. In DPEF, the pipe moves continuously throughout the UAV’s traversal. The three metrics reported are: Average Time (seconds, time from disturbance start until error stays below 10\% threshold), Max Error (peak percentage error during the transient), and Average Error (mean percentage error over the whole trial). The corresponding numerical results are listed in TABLE \ref{tab:performance}.}
    \label{fig:14}
\end{figure}

\begin{figure}
    \centering
    \includegraphics[width=1\linewidth]{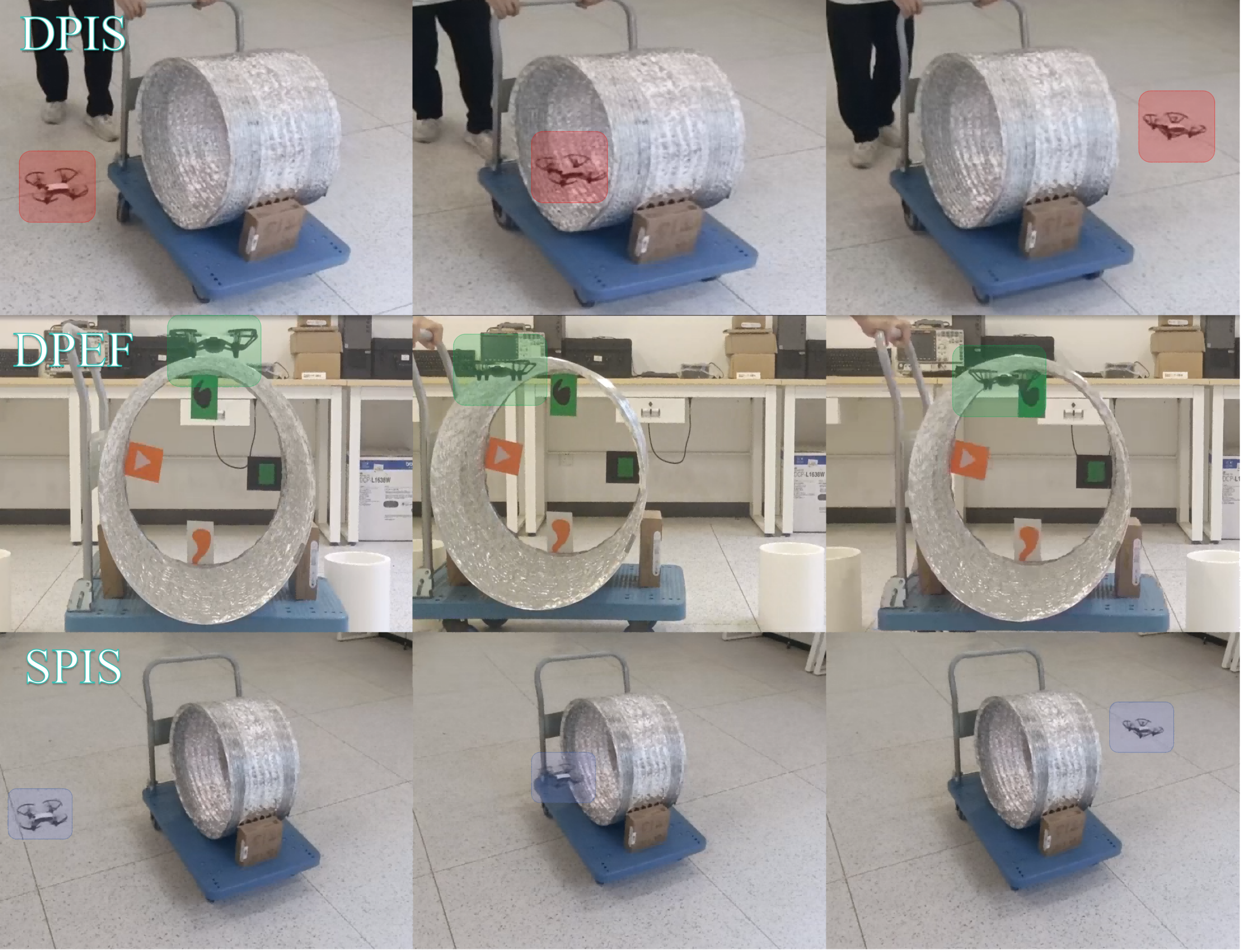}
    \caption{\justifying Evaluating the performance of the MPC-RCAPID method versus the traditional PID method under three testing schemes: DPIS, DPEF, and SPIS. The figure shows the posture of the drone during tracking under these three testing schemes.}
    \label{fig:13}
\end{figure}

This experiment establishes three test dimensions: (1) Dynamic Pipes Internal Shuttle (DPIS); (2) Dynamic Pipes External Following (DPEF); (3) Static Pipes Internal Shuttle (SPIS), as shown in Fig. \ref{fig:13}. The MPC-RCAPID and the standard PID controller are compared across three experimental scenarios. The experimental results are shown in Fig. \ref{fig:14} and TABLE \ref{tab:performance}. The results demonstrate that the MPC-RCAPID control method achieves a more timely tracking response to the moving pipe. 
This method outperforms the standard PID controller in both maximum and average tracking error. Specifically, the maximum tracking error is reduced by 50.63\%, 17.17\%, and 17.45\% in DPIS, DPEF, and SPIS. The average tracking error is reduced by 55.52\%, 26.72\%, and 26.52\%. The smaller tracking errors effectively increase the error tolerance space for UAVs flying in the pipe. This reduction significantly decreases the risk of collision with the inner wall of the pipe. Consequently, the success rate of the dynamic pipe tracking task is ensured. Based on the results of simulation experiments and physical machine experiments, it is concluded that the VisGeo-PipeTrack framework  provides reliable  support for UAVs to efficiently complete tracking and following tasks in a dynamic pipe environment.

\begin{table}
    \centering
    \renewcommand{\arraystretch}{1}
    \caption{Comparison of Tello UAV Experimental Results}
    \label{tab:performance}
    \begin{tabular}{>{\bfseries}c
                   >{\centering\arraybackslash}m{1.8cm}  
                   >{\centering\arraybackslash}m{1.2cm}  
                   >{\centering\arraybackslash}m{1.2cm}    
                   >{\centering\arraybackslash}m{1.4cm}} 
    \toprule
    \textbf{Categories} & \textbf{Scheme} & \textbf{Average Time} & \textbf{Max Error} & \textbf{Average Error} \\
    \midrule
    \multirow{2}{*}{DPIS} & MPC-RCAPID & 7.41s & 17.34\% & 6.73\% \\
                           & PID & 7.35s & 35.12\% & 15.13\% \\
    \midrule
    \multirow{2}{*}{DPEF} & MPC-RCAPID & 106.19s & 36.27\% & 5.87\% \\
                           & PID & 121.80s & 43.79\% & 8.01\% \\
    \midrule
    \multirow{2}{*}{SPIS} & MPC-RCAPID & 3.82s & 9.84\% & 7.26\% \\
                           & PID & 7.82s & 11.92\% & 9.88\% \\
    \bottomrule
    \end{tabular}
\end{table}

\section{Conclusion}
This study proposes a visual geometry-based dynamic pipeline tracking servo control framework for unmanned aerial vehicles (UAVs) named VisGeo-PipeTrack, which applies to both internal traversal and external tracking tasks in dynamic pipe environments. To address challenges posed by dynamic changes within confined pipeline spaces, the framework integrates lightweight visual geometry localization methods with MPC-RCAPID control strategies, enabling UAVs to execute swift and stable attitude adjustments. The framework's effectiveness is validated through AirSim simulations and experiments using the Tello UAV platform.

Future research will concentrate on two main avenues to enhance the framework’s practicality and scalability. Firstly, the framework’s application will be expanded to irregular pipes and those with multiple internal obstacles. The label placement strategy and geometric positioning algorithm will be optimized to improve environmental adaptability. Secondly, the YOLOv5 model and the MPC-RCAPID algorithm will be optimized for lightweight implementation. This will facilitate the integration of perception and control in the UAV’s airborne embedded system. It is helpful to reduce reliance on the ground station and enhance the autonomy of UAV pipes tracking.

\bibliography{references}
\bibliographystyle{IEEEtran}

\vfill

\end{document}